\documentclass{article}

     \usepackage[final]{neurips_2019}

\usepackage[utf8]{inputenc} 
\usepackage[T1]{fontenc}    
\usepackage{hyperref}       
\usepackage{url}            
\usepackage{booktabs}       
\usepackage{amsfonts}       
\usepackage{nicefrac}       
\usepackage{microtype}      
\usepackage{graphicx, float}

\usepackage{fancyhdr}
\usepackage[normalem]{ulem}
\usepackage{comment}
\usepackage{color}
\usepackage{amssymb}
\usepackage{pifont}
\usepackage{algorithm}
\usepackage{algorithmic}
\usepackage{tikz}
\usetikzlibrary{shapes.geometric, arrows}
\usepackage[labelfont=bf, textfont=bf,font=small]{caption}
\usepackage{graphicx}
\usepackage{subcaption}

\title{
DAWSON: A Domain Adaptive \\ 
Few Shot Generation Framework \\ 
{\normalsize  Music MATINEE: Domain Adaptive 
Music Generation via Meta Learning 
}
}

\author{%
Weixin Liang, Zixuan Liu, Can Liu\thanks{Listed in the order of SUID. 
Work done as a course project for CS 236 Deep Generative Models at Stanford University 
https://deepgenerativemodels.github.io/postersess.html
} \\
  Stanford University, CA \\
}
\begin{document}

\tikzstyle{startstop} = [rectangle, rounded corners, minimum width=2cm, text width=2cm, minimum height=2cm, text centered, node distance=8cm, draw=black, fill=red!30]
\tikzstyle{io} = [rectangle, rounded corners, minimum width=3cm, minimum height=1cm, text centered, text width=1cm, node distance=3cm, draw=black, fill=blue!30]
\tikzstyle{process} = [rectangle, minimum width=3cm, minimum height=1cm, text width=3cm, text centered, draw=black, fill=orange!30]
\tikzstyle{decision} = [diamond, minimum width=3cm, minimum height=1cm, text centered, text width=3cm, draw=black, fill=green!30]
\tikzstyle{arrow} = [thick,->,>=stealth]

\newcommand{\weixin}[1]{[{\color{blue}WX: #1}]}

%


\newcommand{\modelabbrv}[1]{{{{DAWSON}}}{#1}}
\newcommand{\modelname}[1]{{{{\underline{D}omain \underline{A}daptive Fe\underline{w} \underline{S}h\underline{o}t Ge\underline{n}eration Framework For GANs}}}{#1}}

\maketitle

\begin{abstract}
Training a Generative Adversarial Networks (GAN) 
for a new domain 
from scratch 
requires an enormous amount of training data 
and 
days of training time. 
To this end, 
we propose \modelabbrv, a \modelname{} 
based on meta-learning.~\footnote{Our model was originally called as Music MATINEE. As we make our progress, we find that our plug-and-play framework for general few-shot generation with GANs might also be interesting. 
Therefore, we re-position our work as \modelabbrv, a \modelname. 
}  
A major challenge of applying meta-learning on GANs 
is to obtain gradients for the generator from evaluating it on development sets 
due to the likelihood-free nature of GANs. 
To address this challenge, 
We propose an alternative GAN training procedure 
that naturally combines 
the two-step training procedure of GANs 
and the two-step training procedure of meta-learning algorithms. 
\modelabbrv{} is a plug-and-play framework 
that supports a broad family of meta-learning algorithms and 
various GANs with architectural-variants. 
Based on \modelabbrv, 
We also propose MUSIC MATINEE, 
which is the first few-shot music generation model. 
Our experiments show that 
MUSIC MATINEE could quickly adapt to new domains with only tens of songs from the target domains. 
We also show that 
\modelabbrv{} can learn to generate new digit with only four samples in the MNIST dataset. 
We release source codes implementation of \modelabbrv{} in both PyTorch and Tensorflow, 
generated music samples on two genres
and the lightning video. 
\end{abstract}

\section{Introduction}


Generative adversarial networks (GANs) have achieved promising results in various domains such as music generation~\cite{musegan}, 
computer vision~\cite{ganCV}, natural language processing~\cite{ganNLP1}, time series synthesis~\cite{ganTimeSeries}, 
semantic
segmentation~\cite{GANsemanticSegmentation} etc. 
However, 
GANs are notoriously hard to train, especially for high-dimensional data like music and image. 
For example, 
training a music generation model 
requires at least millions seconds of training music 
and tens of hours of training time. 
Without enough training data, 
there is also a risk of overfitting for models with high capacity. 
Therefore, current GANs are not well-suited for rapid learning in real-world applications where data acquisition is expensive or fast adaptation to new data
is required. A fast adaptation of the trained model with limited number of training data in the target domain would be desirable.

\begin{figure}[tb]
\centering
\includegraphics[width=0.6\textwidth]{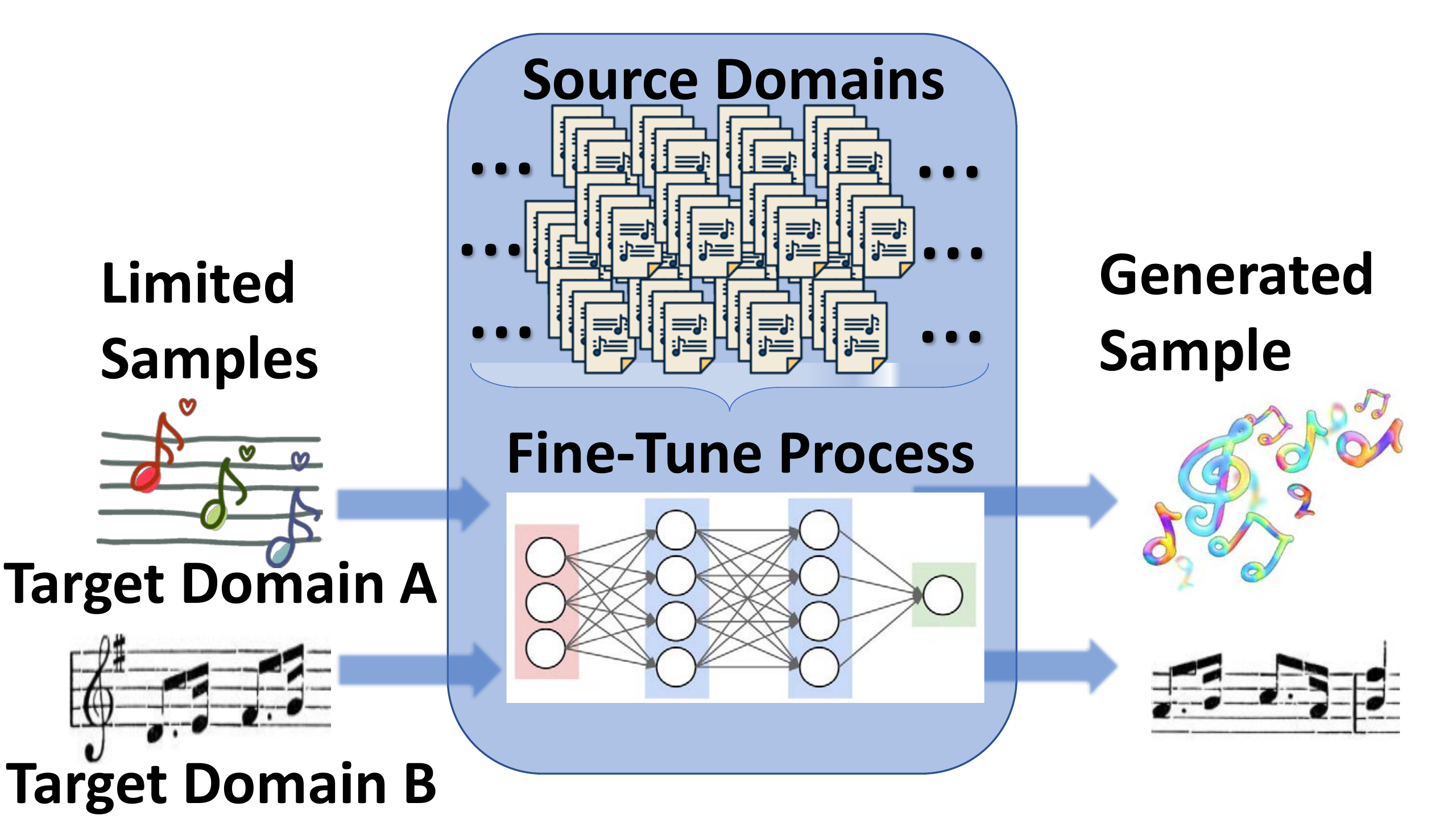} 
\caption{Few Shot Generation. The goal of few-shot generation is to 
provide a pre-trained model during an offline training phrase. 
The pre-trained model is expected to generate high-quality samples with only a limited number of samples in the target domains provided. 
}
\label{fig:fewshot}
\end{figure}

In this paper, we take on the challenge of the few-shot generation problem, which is only  sporadically  explored~\cite{fewshotGen1,fewshotGen2,fewshotGen3,figr}. As illustrated in Figure~\ref{fig:fewshot}, 
the goal of few-shot generation is to 
build an internal representation 
by pre-training on multiple source domains. 
During testing, a limited number of samples in the target domains are provided. 
The pre-trained model is expected to generate high-quality samples with only a limited number of samples in the target domains provided. 
Closely related to few-shot generation, 
few-shot classification aims to train a  classifier 
that could be adapted to unseen classes in the training data given only a few examples of each of these classes. 
It has been shown that naive approaches such as re-training the model on the new data, would severely over-fit~\cite{DBLP:conf/nips/SnellSZ17}. 
Recently, meta-learning algorithms like MAML~\cite{maml} 
has shown promising results in few-shot classification and is widely adopted. 
The key idea of the meta-learning algorithm family of MAML is 
to train the model’s
initial parameters such that the model has maximal performance on a new task after the parameters have been updated through one or more gradient steps computed with 
a small amount of data from that new task~\cite{metaDialog}. 
However, the likelihood-free nature of GANs impose a fundamental challenge in combining it with the meta-learning algorithms. 
To the best of our knowledge, 
no previous work integrates MAML with any GANs for generation tasks. 

We propose \modelabbrv, a \modelname. 
We address the challenges of incorporating GANs and meta-learning algorithms 
in a general way so that \modelabbrv{} could support a broad family of meta-learning algorithms and 
various GANs with architectural-variants. 
In meta-learning algorithms like MAML, 
a key step is to 
evaluate the gradients of the model on development sets. 
However, due to the likelihood-free nature of GANs, 
there is no straight-forward way of obtain gradients for the generator from evaluating it on development sets. 
To address this challenge, 
We propose an alternative GAN training procedure 
that naturally combines 
the two-step training procedure of GANs 
and the two-step training procedure of meta-learning algorithms. 
\modelabbrv{} is a plug-and-play framework 
that supports a broad family of meta-learning algorithms and 
various GANs with architectural-variants.

Based on the plug-and-play feature of \modelabbrv, 
we also propose the first few-shot music generation model, MUSIC MATINEE. 
We select music domains because 
AI-based music generation has already show high socio-economic impacts. AI music composer startups like Amper and Jukedeck has raised millons of funding~\cite{startups}. 
Since \modelabbrv{} is agnostic to the architecture of GANs, 
we can easily plug the state-of-the-art  multi-track music generation model  MuseGAN~\cite{musegan} 
into \modelabbrv. 
Our qualitative experiments show that 
MUSIC MATINEE could quickly adapt to new domains with only tens of songs from the target domains. 
In addition, 
on the MNIST dataset, 
we also show that 
\modelabbrv{} can learn to generate new digit with only four samples in target domain (digit '9') within a short meta-training time and fine-turning time. Such property is extremely useful for the community to explore the variants of \modelabbrv{} in fast speed to push the frontier of few-shot generation for GANs.

In summary, we make the following contributions:  

\begin{itemize}
    \item \textbf{Algorithmic contributions} We propose \modelabbrv, a \modelname. 
    We address the fundamental gap between GANs' likelihood-free training and the key meta-learning step to derive gradients from the process of evaluating the model on separate development sets by proposing an alternative GAN training procedure. 
    \modelabbrv{} is a plug-and-play framework 
    that flexibly supports a broad family of meta-learning algorithms and 
    various GANs with architectural-variants. 
    
    \item \textbf{Domain contributions} 
    Music generation has already show high socio-economic impacts~\cite{startups}. 
    We propose the first few-shot music generation model, MUSIC MATINEE based on the plug-and-play feature of \modelabbrv. MUSIC MATINEE could quickly adapt to new domains with only tens of songs from the target domains.

    \item \textbf{Experimental contributions} 
    To encourage a fast exploration of few shot generation for GANs in the community, 
    we release the implementation of \modelabbrv{} in both pyTorch and Tensorflow in image few-shot generation and music few-shot generation respectively. 
    We also release a lightning video with generated music samples. 
    In case of technical issues of opening the links, we list our urls in plain text here: 
    \begin{itemize}
        \item 
        Lightning video with generated samples In Youtube \newline 
        \href{https://www.youtube.com/watch?v=t65C0iWOe_8}{https://www.youtube.com/watch?v=t65C0iWOe\_8}   
        \item PyTorch and Tensorflow Implementation of  \modelabbrv{} In Github: \newline 
        \href{https://github.com/LC1905/musegan/}{https://github.com/LC1905/musegan/} 
        
    \end{itemize}
\end{itemize}

\section{Related Work}


\noindent \textbf{Music Generation: } 
To the best of our knowledge, we are the first to propose few-shot music generation. 
Prior research efforts on music generation domain primarily focus on exploring various neural network architectures. 
One line of the research leverages sequential generative models such as Hidden Markov Models~\cite{nips2004hmm}, Long Short Term Memory networks~\cite{lstm2002music,lstm2002Bbachbot,lstm2018performancegeneration} 
, bidirectional LSTMs~\cite{deepbach2017icml}, hierarchical LSTMs~\cite{hred2017music,hred2018music} and transformer~\cite{musictransformer}. 
These methods 
sequential generative methods like HMM~\cite{nips2004hmm}, LSTM~\cite{deepbach2017icml} or transformer~\cite{musictransformer}
achieved promising results in generating single-track monophonic music, but they had limited power when it comes to multi-track polyphonic music. This is chiefly because modeling only the temporal structure of polyphonic music fails to take into consideration its other important properties such as harmonic structure, rhythmic structure, and the interdependency between different musical instruments.
On the other end of the spectrum, 
some formulate music as a piano-roll-like 2D representation. 
At each time step, the music is represented as a multi-hot pitch vector. 
Since Piano-rolls are image-like,
some try to model it with CNNs trained either as generative adversarial networks~\cite{musegan} 
or as orderless NADEs~\cite{cnn2017gibbs}. 
To better exploit these ``spatial" structures, some researchers represent multi-track music as image-like pianorolls and model it with CNNs-based generative adversarial networks~\cite{musegan} 
or orderless NADEs~\cite{cnn2017gibbs}. 
We follow the later modeling approach  as we plan to support meta-learning on both single-track monophonic and multi-track music. 
Different from previous work, 
Music MATINEE is the first few-shot music generation model. 

\noindent \textbf{Few Shot Generation} 
Few-shot generation is a challenging problem which is only  sporadically  explored~\cite{fewshotGen1,fewshotGen2,fewshotGen3,figr}. 
\cite{fewshotFirst} applies Bayesian Program Learning to learn simple concepts like pen stroke and combine the concepts hierarchically to generate new images. 
\cite{fewshotGen2} 
models the problem with sequential generative models and augments VAE with an attention module. 
Matching network~\cite{fewshotGen1}, 
variable inference for memory addressing~\cite{VariationalMemoryAddressing} 
and gated linear networks~\cite{gatedlinearNet} have also been proposed. 
\cite{fewshotGen1} employs a memory-assisted matching network to achieve few-shot image generation. 
Sample Efficient Adaptive Text-To-Speech (TTS) ~\cite{tts} learns a user embedding to achieve few-shot adoption. 
\cite{fewShotDensity} extended Pixel-CNN with neural attention for few-shot auto-regressive
density modeling. 
Recently, meta-learning algorithms like MAML~\cite{maml} 
has shown promising results in few-shot classification and is widely adopted. 
However, there is 
a fundamental gap between GANs' likelihood-free training and the key meta-learning step to derive gradients from the process of evaluating the model on separate development sets. 
To the best of our knowledge, 
no previous work integrates MAML with any GANs. 
We are only aware of limited previous work~\cite{figr,metalGAN} which only support a specific kind of meta-learning algorithm (Reptile) with vanilla GAN architecture. In contrast, we support a broad family of MAML based meta-learning algorithm and a broad family of GANs with architectural variants.

\section{Problem Statement}

\subsection{Notations and Terminologies}
Before proceeding, we first clarify some terminologies that will be used throughout the discussion of our meta-learning algorithms. Similar to meta-learning in classification, meta-learning for generative models is built upon the following concepts: \textbf{class}, \textbf{domain}, \textbf{task}. 

Here a \textbf{class} specifies a categorization of the data based on some similarity metric, and the classes can be disjoint or overlapping. For example, in the scenario of natural image generation, $Cat$ and $Dog$ can be considered as two classes that are mutually exclusive, while $Cat$ and $Pet$ are also valid classes, despite that a significant portion of them intersects. Still, as we will discuss later, decent performance of meta-learning algorithms require well-separated classes, so we can think of the classes here as ``almost disjoint". We use $C = \{c_1, c_2, \cdots, c_n...\}$ to denote the collection of all classes in our dataset. 

In terms of samples, we use $D$ to denote the entire dataset and $p_{data}(\mathbf{x})$ its distribution. Furthermore, we define a \textbf{domain} of class $c_i$, denoted by $D_{c_i}$, as the collection of samples in class $c_i$. In other words,
\[D_{c_i} = \{\mathbf{x} \in D~|~class(\mathbf{x}) = c_i\}\]
And we use $p_{c_i}(\mathbf{x})$ to denote the distribution of this domain.

Finally, a \textbf{task} is a tuple consisting of a collection of $r$ classes $C_\mathcal{T} = \{c_{\mathcal{T}_1}, \cdots c_{\mathcal{T}_r}\}$, a distribution $q_\mathcal{T}(\mathbf{x})$ of samples from domain $D_{c_{\mathcal{T}_1}} \cdots D_{c_{\mathcal{T}_r}}$, and a task-specific loss function $L_\mathcal{T}$, i.e.

\[\mathcal{T} = \{C_\mathcal{T}, q_\mathcal{T}(\mathbf{x}), L_\mathcal{T}\}\]

In the scenario of generative models, a task $\mathcal{T}$ can be interpreted as ``generating samples that look similar to data of class $C_{\mathcal{T}_1}$, $C_{\mathcal{T}_2} \cdots$ or $C_{\mathcal{T}_r}$", and the generation quality is evaluated via loss $L_\mathcal{T}$. Note that it is possible to have $r > 1$, i.e. there are more than one class in the current task that the generative model aims to imitate. 
In addition, we use $p(\mathcal{T})$ to denote the task distribution where we apply our meta-learning algorithms.

\subsection{Meta-Learning for GANs} 
Few-shot meta-learning, when it comes to generative models, aims at training a model that quickly adapts to a different (but related) generation task after being optimized on a few examples from the new domain. To be more precise, a model with $K$-shot learning capacity should be able to generate samples in some class $c \in C_\mathcal{T}$ after being trained on $K$ data points sampled from $q_\mathcal{T}(\mathbf{x})$. In this project, we particularly look at a class of likelihood-free generative models known as Generative Adversarial Networks (GANs), and, in this case, fast adaptation of both the generator and the discriminator are desired. We will now formally formalize the problem of meta-learning for generative models and provide some concrete examples. 

Let $f = \{G_{\theta}, D_{\phi}\}$ be a GAN-based generative model, where $G_{\theta}$ and $D_{\phi}$ denote the generator and the discriminator, respectively. The generator 
\[G_{\theta}: \mathbb{R}^d \to \mathbb{R}^{D}\]
takes a noise $\mathbf{z} \in \mathbb{R}^d$, and maps it to a generated sample $\mathbf{x} = G_{\theta}(\mathbf{z}) \in \mathbb{R}^D$. With a mild abuse of notation, we use $\mathbb{R}^D$ to represent the dimension of the data, where, usually, $d \ll D$. The discriminator 
\[D_{\phi}: \mathbb{R}^D  \to \mathbb{R}^{m}\]

takes a sample $\mathbf{x} \in \mathbb{R}^D$ and outputs a critic of the sample $y = D_{\phi}(\mathbf{x}) \in \mathbb{R}^{m}$. If the discriminator only distinguishes between real and fake samples, then $m = 1$ and $y$ is a scalar describing the probability that the given example is real. If, in addition, the discriminator classifies the examples into different classes, then $m = \mathcal{T}_r+1$, where $\mathcal{T}_r$ is the number of classes in task $\mathcal{T}$. The critic $\mathbf{y} \in \mathbb{R}^{m+1}$ now gives probabilities that the sample is fake, and that it is real and belongs to one of the classes.



Let $T$ be the space of all tasks, which is then partitioned into $T = T_{train} \cup T_{test}$ with $T_{train} \cap T_{test} = \emptyset$.  $T_{train}$ contains tasks used for meta-training and $T_{test}$ those for meta-testing. At the stage of meta-training, at each iteration, we sample a batch of tasks from $T_{train}$ following $p(\mathcal{T})$ and apply meta-update, whose details will be described in later sections. At test time, for K-shot meta-learning, we first draw a test task $\mathcal{T} \in T_{test}$ following $p(\mathcal{T})$. The model is then updated using $K$ samples $\{\mathbf{x}_1, \mathbf{x}_2, \cdots, \mathbf{x}_K\} \sim q_{\mathcal{T}}(\mathbf{x})$ and feedback from $L_{\mathcal{T}}$. Since we are training GANs, the loss function $L_{\mathcal{T}}$ is the objective $L_{\theta, \phi} = \{L_{\theta, \phi}^D, L_{\theta, \phi}^G\}$ used to optimize $f$. The meta-performance is measured by evaluating the samples generated by the updated model in terms of their similarity with the target domain.

\section{Method}

Our end goal is to enable few-shot learning in music generation. To be more precise, we want to incorporate few-shot learning into MuseGAN, a state-of-the-art GAN model for multi-track, polyphonic music generation. In order to accomplish this, we apply MAML-style meta-learning algorithms to GAN-based generative models. The key challenge in applying MAML to GANs, however, is that this line of meta-learning algorithms relies on evaluating the model on a development set to provide a ``gradient" that is later used for meta-update. Unfortunately, due to the likelihood-free nature of GANs,  the feedback from the development set cannot be obtained in a natural and straight-forward way. We will delve deeper into the meta-learning algorithms, and further address this issue as well as our solution in the second part of the following discussion. Before that, we will also briefly discuss the architectures and losses adopted for our GAN models.

\subsection{Generative Adversarial Networks for Few-Shot Learning}

\subsubsection{MuseGAN}
There are some variations in MuseGAN models, and we exploit the most recent version \cite{dong2018convolutional}.
The model consists of a generator, a discriminator, and a refiner that maps the real-value output of the generator into pianoroll binaries. Both the generator and the discriminator are composed of one ``shared" and $M$ ``private" networks. The shared component in the generator captures high-level, inter-track features of the music, while each private network further decodes this high-level representation into the pianoroll of a particular track. The shared and private components in the discriminator work in a similar but reverse manner. The model is trained using WGAN-GP \cite{gulrajani2017improved} loss, which stabilizes the training. Thus our task loss $L_{\mathcal{T}} = \{L_{\theta, \phi}^D, L_{\theta, \phi}^G\}$ is defined as:
\[L_{\theta, \phi}^D = \mathbb{E}_{\mathbf{x} \sim p_{data}} [D_\phi(\mathbf{x})] - \mathbb{E}_{\mathbf{z} \sim N(0, I)} [D_\phi(G_\theta(\mathbf{z}))] + \lambda \mathbb{E}_{\hat{\mathbf{x}} \sim p_{\hat{\mathbf{x}}}} [(||\nabla_{\hat{\mathbf{x}}} D(\hat{\mathbf{x}})||_2 - 1)^2]\]
\[L_{\theta, \phi}^G = -\mathbb{E}_{\mathbf{z} \sim N(0, I)} [D_\phi(G_\theta(\mathbf{z}))]\]

Here $\hat{\mathbf{x}} \sim p_{\hat{\mathbf{x}}}$ is generated by first sampling $\alpha \sim U(0, 1)$, and then interpolating between a batch of real data and a batch of fake data using $\alpha$, i.e.
\[\hat{\mathbf{x}} = \alpha \cdot \mathbf{x}_{real} + (1-\alpha) \cdot \mathbf{x}_{fake}\]

\begin{algorithm}[tbh]
\caption{\modelabbrv{} Meta-Learning Inner Loop for Plugging MAML in: 
}
\begin{algorithmic}
\REQUIRE $\mathcal{T}_i$: A single task
\REQUIRE $G_{\theta}$: Real generator; $D_{\phi}$: Real discriminator
\REQUIRE $G_{\hat{\theta}}$: Cloned generator; $D_{\hat{\phi}}$: Cloned discriminator
\REQUIRE $\alpha, L$: Inner learning rate; Number of inner loop updates.
\STATE Split samples in $\mathcal{T}_i$ into disjoint training and dev sets.
\STATE Draw K training samples $D_{train} =\{\mathbf{x}_1, \cdots, \mathbf{x}_K\} \sim$ $q_{\mathcal{T}_i}$ 
\STATE Draw K dev samples $D_{dev} =\{\mathbf{y}_1, \cdots, \mathbf{y}_K\} \sim$ $q_{\mathcal{T}_i}$ 
    \FOR{$i=1$ \TO $L$}
        \STATE Generate a minibatch of fake samples $\tilde{\mathbf{x}} \gets G_{\hat{\theta}}(\mathbf{z})$ with $\mathbf{z} \sim N(0, I)$
        \STATE Update $D_{\hat{\phi}}$: $\hat{\phi} \gets \hat{\phi} - \alpha \cdot \nabla_{{\hat{\phi}}} L_{\hat{\theta}, \hat{\phi}}^D(D_{train}, \tilde{\mathbf{x}})$
    \ENDFOR
    \STATE Compute ``Gradient" for discriminator: $\nabla D \gets \nabla_{\phi} L_{\hat{\theta}, \hat{\phi}}^D(D_{dev}, \mathbf{x})$
    \STATE Clone $D_{{\hat{\phi}_{dev}}} \gets D_{\phi}$
    \FOR{$i=1$ \TO $L$}
        \STATE Generate a minibatch of fake samples $\tilde{\mathbf{x}} \gets G_{\hat{\theta}}(\mathbf{z})$ with $\mathbf{z} \sim N(0, I)$
        \STATE Update $D_{{\hat{\phi}}_{dev}}$: $\hat{\phi}_{dev} \gets \hat{\phi}_{dev} - \alpha \cdot \nabla_{{\hat{\phi}}} L_{\hat{\theta}, \hat{\phi}_{dev}}^D(D_{dev}, \tilde{\mathbf{x}})$
    \ENDFOR
    \FOR{$i=1$ \TO $L$}
        \STATE Update $G_{\hat{\theta}}$: $\hat{\theta} \gets \hat{\theta} - \alpha \cdot \nabla_{{\hat{\theta}}} L_{\hat{\theta}, \hat{\phi}}(\tilde{\mathbf{x}})$
    \ENDFOR
\STATE Compute ``Gradient" for generator: $\nabla G \gets \nabla_{\phi} L_{\hat{\theta}, \hat{\phi}_{dev}}^G(\tilde{\mathbf{x}})$
\STATE Return $\nabla D$, $\nabla G$
\end{algorithmic}
\label{MetaTrainingMAMLInnerLoop}
\end{algorithm}

\subsubsection{ACGAN}
The problem with MuseGAN, however, is that the discriminator is only designed for binary real/fake classification, which is not optimal for our few-shot generation task that requires domain adaptation. In other words, feedback from the task loss $L_{\mathcal{T}}$ only evaluates how similar the generated samples are to the real data distribution, but not how similar they are to the particular domain distribution that our model is learning from. In order to also measures the  ``likeliness" of the generated samples to a domain, we introduce an ACGAN-like architecture
, where the discriminator is formulated as a multi-class classifier and a Cross Entropy Loss is calculated to measure the samples' similarity to the target domain. However, unlike ACGAN, we refrain from converting the generator into a conditional generator and rely purely on samples of new classes for domain adaptation.

\subsubsection{DAWSON: a General Few Shot Generation Framework For GANs}
We incorporate two model-agnostic meta-learning algorithms into the training procedure of GANs: Model-Agnostic-Meta-Learning (MAML) \cite{maml} and Reptile \cite{reptile}. Both algorithms are first purposed for classification tasks, while the later can be considered as a first-order approximation for MAML \cite{reptile}. The key idea behind both models is to find an internal representation of the data that is highly generalizable to different tasks through a clever initialization of the model parameters. And they are also similar in the sense that, at each iteration of meta-training, a tentative update (e.g. through Stochastic Gradient Descent) will first be applied to a cloned version of the model, a process that we call ``inner loop training". After the inner loop update, a ``gradient" will be sent back and used to optimize the original model, and we call this procedure ``outer loop training" Algo \ref{MetaTrainingOuterLoop}.

The way that the ``gradient" from the inner loop training is computed differs between Reptile and MAML. 
For Reptile, the ``gradient" is defined as the difference between model parameters before and after the inner loop update. So the outer loop update rule is
\[\theta \gets \theta - \beta \nabla := \theta - \beta (\theta' - \theta)\]
where $\theta, \theta'$ denotes the mode parameters before and after the inner loop training, respectively, and $\beta$ is the outer loop learning rate. Applying Reptile to GANs is straightforward \ref{MetaTrainingReptileInnerLoop}, where we alternate between updating the discriminator and the generator using the same method. 

For MAML, samples in the training task are further split into a training and a development splits. The inner loop training updates over samples from the training split, while the ``gradient" is computed by evaluating the updated model on the development split.
\[\theta \gets \theta - \beta \nabla := \theta - \beta \nabla_\theta L_{dev, \theta'}\]

As mentioned earlier, computing this ``gradient" requires an evaluation of the updated model on the development set. When it comes the generator of GANs, however, this seemingly natural statement suddenly lacks an intuitive interpretation, because the loss of the generator is not directly calculated with respect to the training data, but indirectly provided by the discriminator, which is also a parameterized component of the model. In order to address this issue and integrate a MAML-style algorithm into GANs, we modify the inner loop training procedure for the generator \ref{MetaTrainingMAMLInnerLoop}. To draw a parallel with the classic MAML, we adopt a discriminator trained on the training samples to return loss that is supposed to come from the training set, and a discriminator updated on the development samples to provide loss that is supposed to come from development set.

\begin{algorithm}[tbh]
\caption{\modelabbrv{} Meta-Learning Inner Loop for Plugging Reptile in: 
}
\begin{algorithmic}
\REQUIRE $\mathcal{T}_i$: A single task
\REQUIRE $G_{\theta}$: Real generator; $D_{\phi}$: Real discriminator
\REQUIRE $G_{\hat{\theta}}$: Cloned generator; $D_{\hat{\phi}}$: Cloned discriminator
\REQUIRE $\alpha, L$: Inner learning rate; Number of inner loop updates.
\STATE Draw K samples $\{\mathbf{x}_1, \cdots, \mathbf{x}_K\} \sim$ $q_{\mathcal{T}_i}$ 
\FOR{$i=1$ \TO $L$}
    \FOR{$k=1$ \TO $K$}
    \STATE Generate fake sample $\tilde{\mathbf{x}} \gets G_{\hat{\theta}}(\mathbf{z})$ with $\mathbf{z} \sim N(0, I)$
    \STATE Update $D_{\hat{\phi}}$: $\hat{\phi} \gets \hat{\phi} - \alpha \cdot \nabla_{{\hat{\phi}}} L_{\hat{\theta}, \hat{\phi}}^D(\mathbf{x}_k, \tilde{\mathbf{x}})$
    \STATE Update $G_{\hat{\theta}}$: $\hat{\theta} \gets \hat{\theta} - \alpha \cdot \nabla_{{\hat{\theta}}} L_{\hat{\theta}, \hat{\phi}}^G(\tilde{\mathbf{x}})$
    \ENDFOR
\STATE Compute ``Gradient" for discriminator: $\nabla D \gets \phi - \hat{\phi}$
\STATE Compute ``Gradient" for generator: $\nabla G \gets \theta - \hat{\theta}$
\STATE Return $\nabla D$, $\nabla G$
\ENDFOR
\end{algorithmic}
\label{MetaTrainingReptileInnerLoop}
\end{algorithm}

\begin{algorithm}[tbh]
\caption{\modelabbrv: Meta-Learning Outer Loop}
\begin{algorithmic}
\REQUIRE $p(\mathcal{T})$: Task distribution.
\REQUIRE $G_\theta$: Generator; $D_\phi$: Discriminator
\REQUIRE $\alpha, \beta, L$: inner learning rate; outer learning rate; number inner loop updates.
\STATE Randomly initialize $\theta, \phi$
\WHILE{not done}
    \STATE Sample batch of training tasks $T_i \sim p(\mathcal{T})$
    \FOR{$T_i$}
    \STATE Clone $D_{\hat{\phi}} \gets D_{\phi}$
    \STATE Clone $G_{\hat{\theta}} \gets G_{\theta}$
    \STATE $\nabla D$, $\nabla G$ $\gets$ \textbf{Inner Loop}$(D_{\hat{\phi}}, G_{\hat{\theta}}, D_\phi, G_\theta, \alpha, L, \mathcal{T}_i)$
    \STATE Update $\phi \leftarrow \phi - \beta \cdot \nabla D$
    \STATE Update $\theta \leftarrow \theta - \beta \cdot \nabla G$
    \ENDFOR
\ENDWHILE
\end{algorithmic}
\label{MetaTrainingOuterLoop}
\end{algorithm}

\section{Experiment}

\subsection{Overview}
To testify the few-shot generation capability 
of \modelabbrv, 
we evaluate the framework in two both image and music generation settings. 
\begin{itemize}
    \item \textbf{Image Generation}: 
    To enable a fast exploration of \modelabbrv, we perform our experiments on the MNIST dataset. We only provide training data with digits 0 to 8 in meta-learning phrase. 
    We evaluate the few-shot generation ability by only providing four samples (four-shot) in digit 9 and then draw samples from the model. 
    \item \textbf{Multi-track Music Generation}
    Since \modelabbrv{} is a plug-and-play framework 
    that supports GANs with architectural-variants, we plug in MuseGAN and get the first few-shot music generation model, MUSIC MATINEE. 
    We evaluate Music MANTINEE on 
    a augmented music dataset with meta-data from the Lakh MIDI Dataset (which does not have meta-dat) and the Million Song Dataset. 
    We use the meta-data on generes to partition tasks. 
    During evaluation, we provide tens of songs on held-out genres e.g., punk, blues. 
\end{itemize}

\subsection{Music Data Representation and Dataset}

To apply meta-learning, 
we need to specify tasks based on the meta-data of musics. We first introduce the dataset used by  MuseGAN~\cite{dong2017musegan}. 
We then introduce how we migrate to a augmented version of this dataset. 
The origin dataset uxed by MuseGAN contains multi-track piano-roll-sytle representation of rock music derived from Lakh MIDI Dataset. 
The majority of our data is rock music in C key, 4/4 time, with pitches ranging from C1 to B7 ($84$ notes in total). Each piano-roll is a concatenation of $5$ tracks, describing $5$ different musical instruments respectively: bass, drums, guitar, strings, and piano. The data tensor for each bar is therefore of shape $96$ (time step) $\times$ $84$ (note) $\times$ $5$ (track). 

However, we note that the dataset does not have any meta-data. 
Fortunately, we find that there is existing resources linking the Lakh MIDI Dataset (which does not have meta-dat) with the 
\href{http://millionsongdataset.com/}{Million Song Dataset}, 
which provides rich meta-data for each music. With this resource, 
we migrated and analyzed 
a dataset of 21,425 songs with 
rich meta-data (e.g., artists, styles). 


We identified the following useful meta data tags in our data-set along with each music. All this meta-data could be utilized to perform clustering as a baseline method for dividing meta-training tasks. Our high level semantic metadata includes: 
\textit{(1)} 
\textbf{Artist\_style\_tags}: the tags given to the artist, for example, rock, classic, hard rock, etc. The frequency, as well as the weights are also provided. Each song is associated with 1~15 tags. \textit{(2)} 
\textbf{Similar\_artist}: a graph of artist with each artist works as an node. Each node has 100 edges connecting to the most similar artists. Given the tags for each artists, we would explore various graph segmentation algorithm to split the articsts into different clusters. 
We have also identified the several low level semantic tags which could provide auxiliary information for clustering artists. Examples include the key, mode and tempo of the song. 

\begin{figure}[tbh]
\centering
\includegraphics[width=0.8\textwidth]{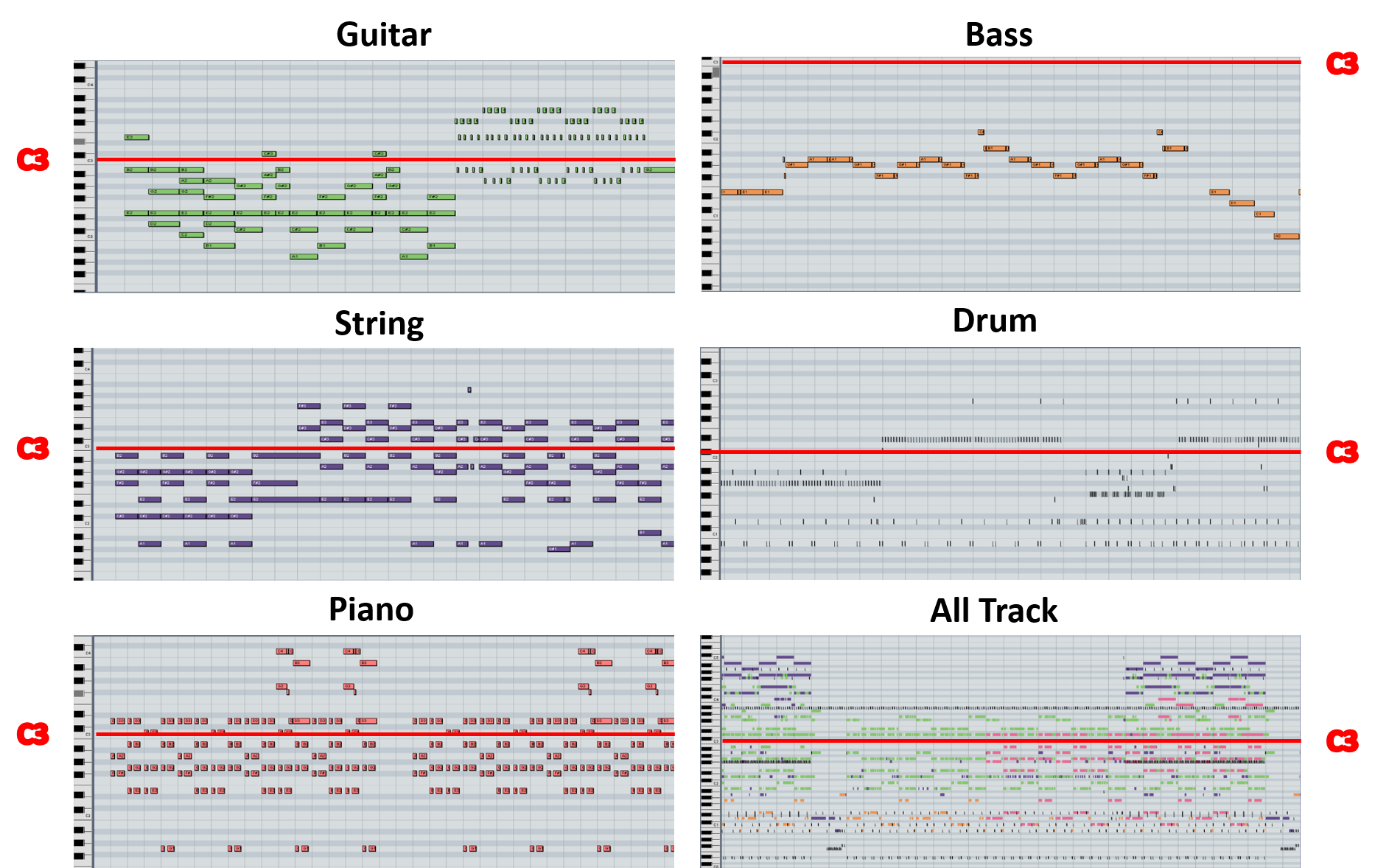} 
\caption{Pianoroll Representation of Multi-Track Music. Each of the first five figures represent an instrument. The x axis represents time and the y axis represents the pitch. At each time step if a pitch is played, the corresponding point in the piannoroll is set to one. Music samples in the training data and the generated music samples along with their pianoroll representations could be viewed in the lightning video. 
}
\label{fig:pianorollraw}
\end{figure}

\subsection{Metrics}
We will evaluate our generative model's potential in few-shot learning from three perspectives: 
\begin{enumerate}
    \item \textbf{Sample Efficiency:} 
    The required amount of training data in the target domain when it is transferred to generate a new type of music/image. This is the fundamental goal of few-shot learning. Here we explore a very extreme case where only four images in the target domain are provided. 
    
    \item \textbf{Generation Quality:} the overall quality of the produced examples. Quantifying the quality of a piece of music is a nontrivial task. This challenge is further complicated by two characteristics of our task: (1) generating music with GAN, which is a likelihood-free model, prevents us from directly querying the Negative Log-Likelihood (NLL), which is a common metric used in other music generation models \cite{musictransformer}; (2) when evaluating the quality of the produced music, we have an additional task of examining how close the generated examples are to the desired type of music, but this similarity is a very subjective notion.
    \item \textbf{Meta-Training Time:} Although the few-shot generation focus more on the fine-tuning stage by definition, we empirically find that meta-training requires 10X or even 100X training time compared the direct training. Therefore, we view the meta-training time as another bottleneck to hinder the exploration of the few-shot generation problem.  
\end{enumerate}

We deliver our quantitative and Qualitative results as well as our analysis in the following.

\subsection{Results and Analysis}

\begin{figure}[htb]
\centering
\centering
    \begin{subfigure}[b]{0.35\textwidth}
        \includegraphics[width=\textwidth]{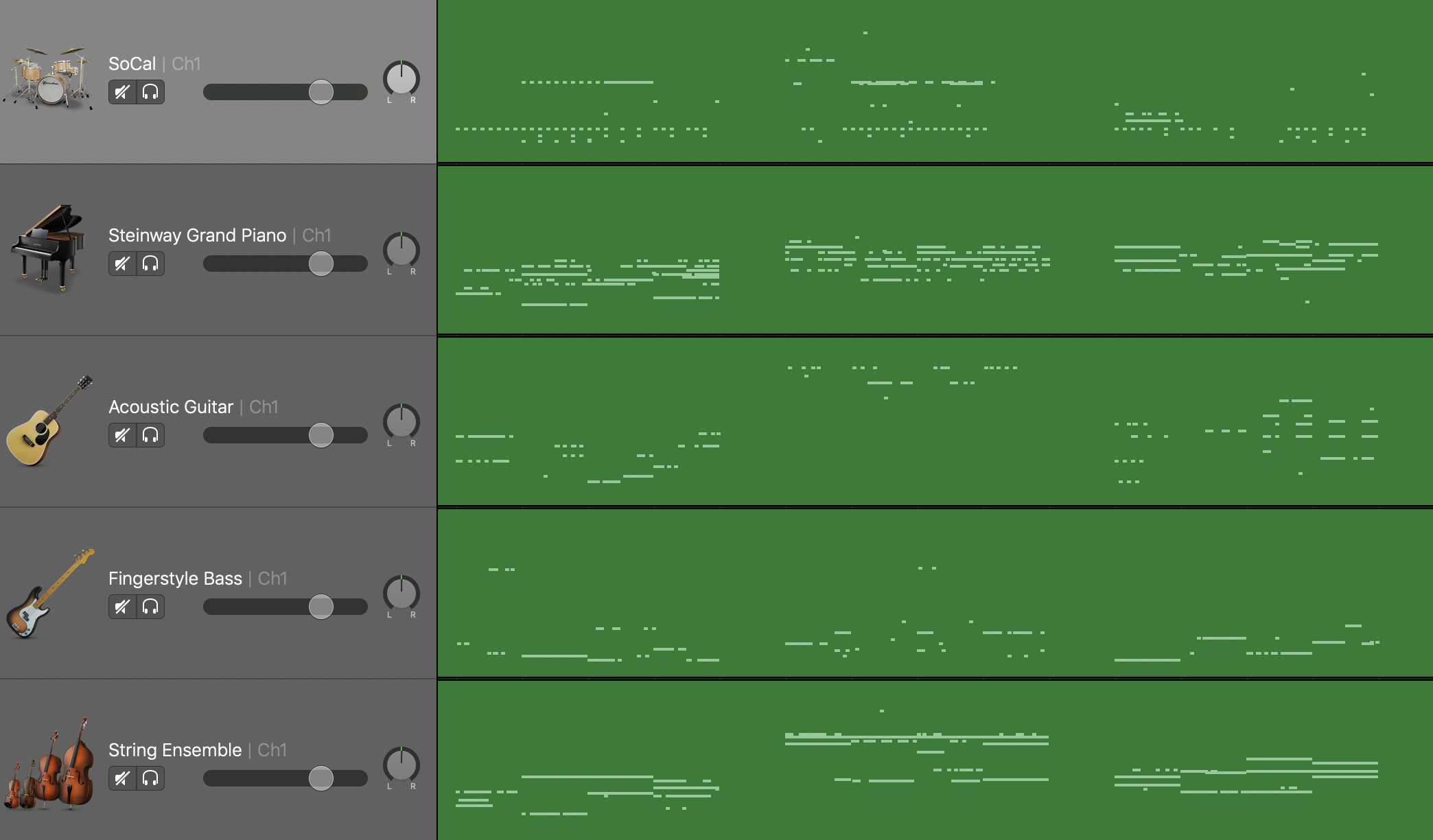}
    \end{subfigure}
    ~ 
    \begin{subfigure}[b]{0.45\textwidth}
        \includegraphics[width=\textwidth]{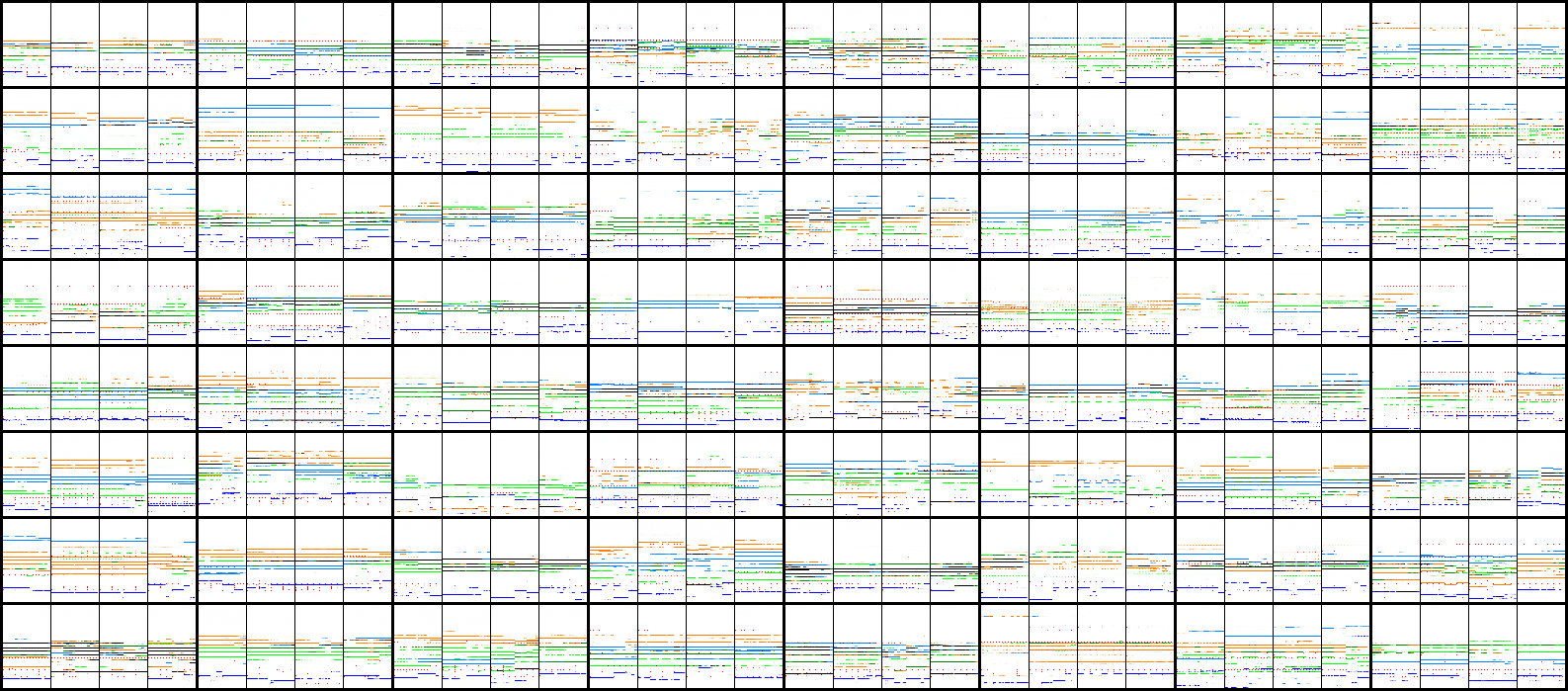}
    \end{subfigure}
\caption{Pianoroll Representation of Generated Musics}
\label{fig:pianoroll}
\end{figure}

The first take-away is that 
\modelabbrv{} is a powerful few shot generation framework for GANs. 
We urge the readers to listen to the generated samples (in blues and punk) attached in our lightning video in YouTube. 
Figure~\ref{fig:pianoroll} also shows a graphical representation of the generated musics. 
We delay the quantitative evaluation of the generated music samples to future work since 
we find that the evaluation for music generation is still an open challenge and there is no straightforward and wide-accepted way 
to evaluate the generated music 
except for human evaluation. We therefore provide more quantitative results in few-shot image generation.

\begin{figure}[htb]
\centering
\includegraphics[width=0.9\linewidth]{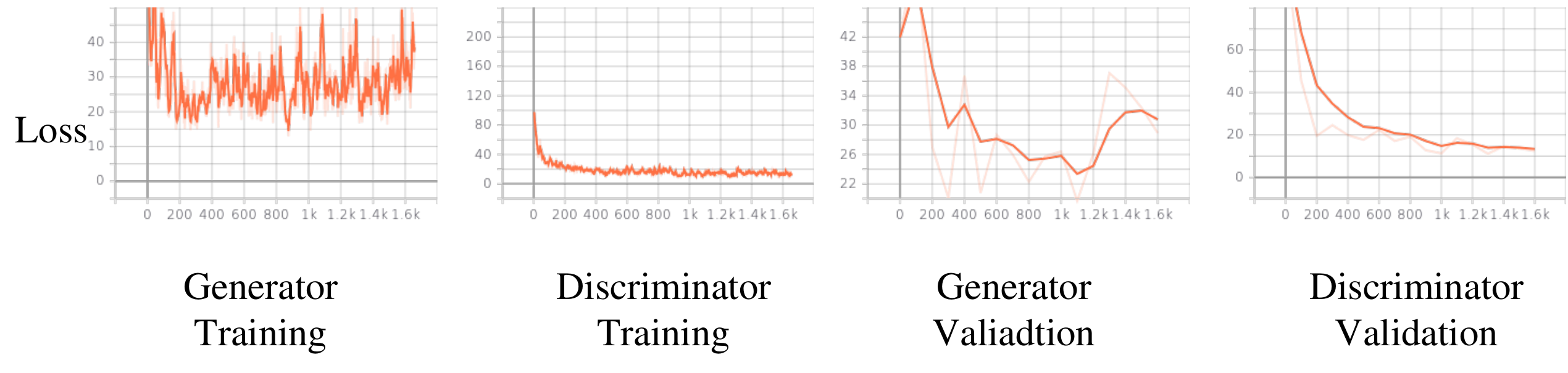} 
\caption{The Loss Curve of \modelabbrv{} on MNIST data-set. The meta-training only takes tens of minutes (1000 epochs) to provide a  good pre-trained model that support few-shot generation on digit '9'.
}
\label{fig:mnistLoss}
\end{figure}

\begin{figure}[htb]
\centering
\includegraphics[width=0.5\linewidth]{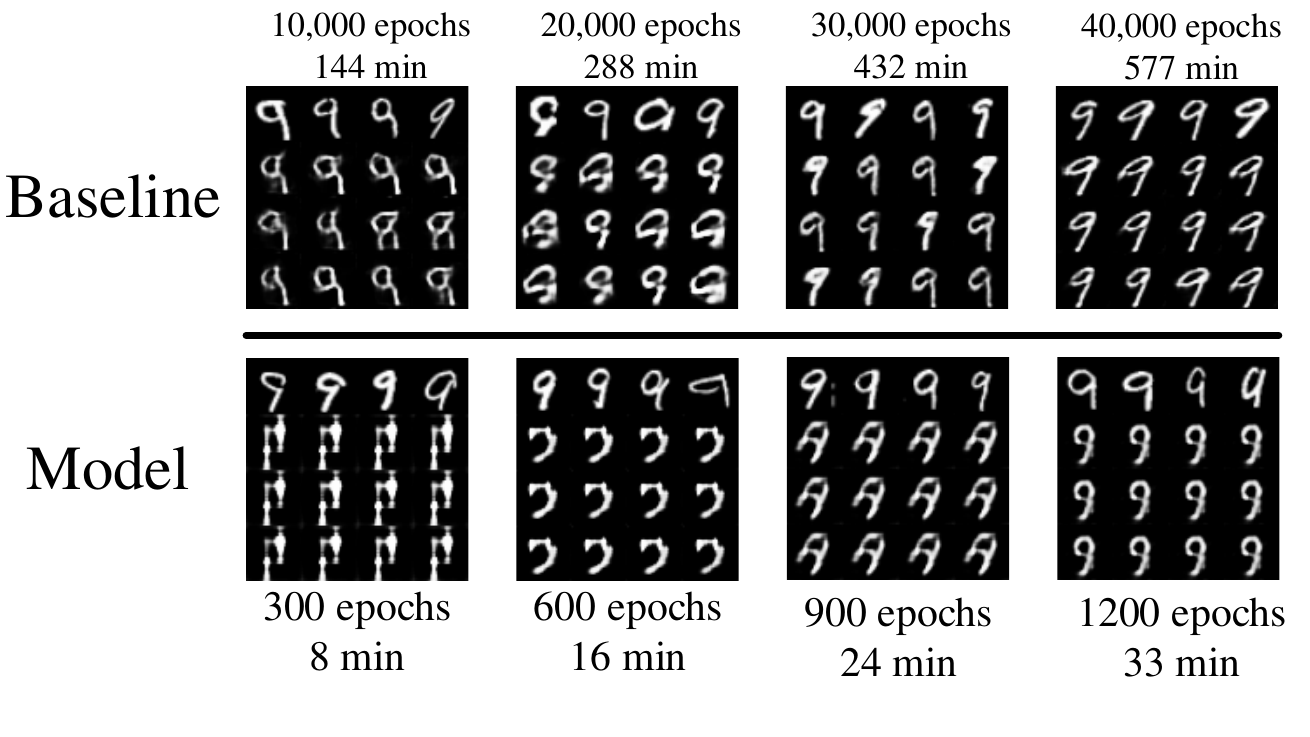} 
\caption{Baseline v.s. Model 1: 4-shot MNIST} 
\label{fig:mnistexample}
\end{figure}

The second take-away is that the flexibility provided by 
the plug-and-play feature of \modelabbrv{} allows 
\modelabbrv{ } to take full advantage of GANs with architectural variants during meta-training. More specifically, 
in MNIST few-shot image generation, 
we show that \modelabbrv{ } could 
obtain better results within 
a significantly faster meta-training time 
by using AC-GAN. 
We compare using AC-GAN and a GAN with a classifier that only tells real and fake. 
As shown in Table~\ref{models_ourdata1} and Figure~\ref{fig:mnistexample}: Adding a classification layer to the discriminator significantly improves the efficiency of the meta-learning algorithms, as well as the sharpness and diversity of the generated images. 

\begin{table}[htb]
\small
\begin{center}
\begin{tabular}{rllll}
\cmidrule[\heavyrulewidth]{1-5}
{Model} & Sharpness 
 & Diversity & Epochs & Meta-Training Time (min)\\  
\cmidrule{1-5}
{Baseline} & 0.324
 & 7.32 & 40,000 & 577 \\  
{Model} & 0.332 & 8.61 & 1,200 & 33 \\  
\cmidrule{1-5}
{MNIST Dataset} & 0.986
 & 9.98 & - & -  \\  
\cmidrule[\heavyrulewidth]{1-5}
\end{tabular}
\caption{Baseline v.s. Model: Sharpness and Diversity}
\label{models_ourdata1}
\end{center}
\end{table}

The third take-away is that 
\modelabbrv{ } also allows the searching in the meta-learning algorithm space to optimize the few-shot generation learning for GANs in terms of different goals. 
We explore MAML~\cite{maml} and Reptile~\cite{reptile}, which is a first-order approximation and an accelerated version of MAML. 
We empirically find that Reptile-style algorithms are easier to train than MAML-style algorithms for generative models. 
Our observation aligns with our expectation 
because (1) \textbf{Reptile bypasses the difficulty of updating generators in MAML} (2) the gradient update is simpler.

The forth take-away is about the future direction of few-shot music generation. We summarize the lessons we learnt on exploring few-shot music generation. We observe that 
even though the performance of the model on MNIST dataset is decent, training a similar meta-learning model on music data is hard. 
In addition to the reason that dimensionality of music data is much higher, 
we suspect that a more important reason is that music data do not have well-defined definitions of similarity and ``tasks".

\section{Conclusion}

In this paper, we incorporate meta-learning with GANs to tackle few shot generation problem. 
More specifically, 
we propose \modelabbrv, a \modelname{} based on meta-learning. 
We address the fundamental gap between GANs' likelihood-free training and the key meta-learning step to derive gradients from the process of evaluating the model on separate development sets by proposing an alternative GAN training procedure. 
\modelabbrv{} is a plug-and-play framework 
that flexibly supports a broad family of meta-learning algorithms and various GANs with architectural-variants. 
Based on the plug-and-play feature of \modelabbrv, 
we propose the first few-shot music generation model, 
MUSIC MATINEE. 
Our experiments in both music and image shows that 
\modelabbrv{} can quickly learn to  
generate high-quality samples 
with limited training data in target domains. 
In the future, 
we plan to add quantitative experiments on more domains. 
In addition, we also want to provide some theoretical 
analysis for \modelabbrv.

\section*{Acknowledgement}
This work was done as a course project for CS 236 Deep Generative Models at Stanford University. 
We would like to thank our TA Kristy Choi for the inpiring discussion and project guidance. We would also like to thank our instructors 
Stefano Ermon
and  
Aditya Grover.

\bibliographystyle{nips}
\bibliography{references}

\end{document}